\documentclass{article}

\PassOptionsToPackage{numbers, compress}{natbib}

\usepackage[preprint]{nips_2018}
\usepackage[utf8]{inputenc} 
\usepackage[T1]{fontenc}    
\usepackage{hyperref}       
\usepackage{url}            
\usepackage{booktabs}       
\usepackage{amsfonts}       
\usepackage{nicefrac}       
\usepackage{microtype}      
\usepackage{times}
\usepackage{epsfig}
\usepackage{graphicx}
\usepackage{amsmath}
\usepackage{amssymb}
\usepackage{algorithm}
\usepackage{algorithmic}
\usepackage{mathrsfs}
\usepackage{array}
\usepackage{bm}
\usepackage{multirow}
\usepackage{wrapfig}

\begin{document}

\title{Dense Multimodal Fusion for Hierarchically Joint Representation}

\author{Di~Hu\\
        School of Computer Science\\
        Northwestern Polytechnical University\\
        \And
        Feiping Nie\\
        School of Computer Science\\
        Northwestern Polytechnical University\\
        \And
        Xuelong~Li\\
        Chinese Academy of Sciences
}

\maketitle
\begin{abstract}
Multiple modalities can provide more valuable information than single one by describing the same contents in various ways.
Hence, it is highly expected to learn effective joint representation by fusing the features of different modalities.
However, previous methods mainly focus on fusing the shallow features or high-level representations generated by unimodal deep networks, which only capture part of the hierarchical correlations across modalities.
In this paper, we propose to densely integrate the representations by greedily stacking multiple shared layers between different modality-specific networks, which is named as \emph{Dense Multimodal Fusion} (DMF).
The joint representations in different shared layers can capture the correlations in different levels, and the connection between shared layers also provides an efficient way to learn the dependence among hierarchical correlations.
These two properties jointly contribute to the multiple learning paths in DMF, which results in \textbf{faster convergence}, \textbf{lower training loss}, and \textbf{better performance}.
We evaluate our model on three typical multimodal learning tasks, including audiovisual speech recognition, cross-modal retrieval, and multimodal classification.
The noticeable performance in the experiments demonstrates that our model can learn more effective joint representation.
\end{abstract}

\section{Introduction}
The same contents or events can be described in multiple kinds of modalities in the real world. That is, the verbal, vocal, and visual modality can be jointly used for expression in different scenarios.
For example, the seen flying plane is always accompanied by loud roar (vocal+visual), speech signal can be expressed by audio and lip movements (vocal+visual), image and caption can jointly describe the concrete contents (verbal+visual), and language can be recorded in both text and audio (verbal+vocal).
As these modalities actually describe the same contents, they are closely related~\cite{sohn2014improved}. Sometimes, one of them can also provide complementary information for the other one. Considering that visual modality is free of audio noise, it can provide efficient information for speech recognition in the noisy environment~\cite{hu2016temporal}.
Hence, multiple modalities can jointly provide more valuable information than single one, and there have been many works over the years making use of multimodal data for specific tasks, such as \emph{Audiovisual Speech Recognition} (AVSR)~\cite{potamianos2004audio}, image-text classification~\cite{srivastava2012multimodal}, and cross-modal retrieval~\cite{wang2016comprehensive}.

Although these works benefit from the valuable multimodal data, different modalities take diverse representations and statistical properties. For example, audio data is usually represented with spectrogram or Mel-Frequency Cepstrum Coefficient in real-value, while textual data is represented using sparse one-hot or word-count vector in discrete value.
These different representations make it difficult to capture the complex correlation across modalities \cite{srivastava2012multimodal}. Fortunately, as these modalities are used to describe the same contents, they should share similar patterns to some extent.
Recently, deep learning methods have shown their effectiveness in generating useful feature representation~\cite{salakhutdinov2009semantic,salakhutdinov2009deep}. Some works, such as \emph{Multimodal Deep Autoencoder} (MDAE) \cite{ngiam2011multimodal} and \emph{Multimodal Deep Boltzmann Machine} (MDBM) \cite{srivastava2012multimodal}, consider that the extracted high-level representations of different modalities are semantically correlated. Hence, they propose to learn a kind of joint representation across the top layers of modality-specific networks. The motivation beyond this strategy is that they assume the high-level representations contain sufficient semantic information and the shared patterns across modalities exist in the semantic level. However, there remain two open questions about such strategy. First, if the high-level representations of each modality can provide sufficient information to capture the complex correlation across modalities, especially when the input data are hand-craft features (e.g, SIFT descriptor, simple word-count vector). Second, if the shared patterns only exist in the semantic level or the representation in specific single layer? In other words, if it is suitable to perform the fusion across modalities for only once?

Actually, the fusion across the high-level representations works like the classical late fusion that fuses the semantic concepts from unimodal features~\cite{snoek2005early}.
Compared with other fusion strategies (e.g., early fusion), the late fusion can only capture the correlation in the semantic level but fail to exploit other kinds of correlations, such as the covariation in the early feature level~\cite{shivappa2010audiovisual,katsaggelos2015audiovisual}, the hierarchical supervision throughout the whole network~\cite{feng2014cross}.
Meanwhile, cognitive researchers have also verified that the multisensory integration exists in not only the temporal lobe but also parietal and frontal lobe~\cite{stein1993merging,holmes2005multisensory}, which indicates that the integration occurs in the different stages of information processing in our brain.
Therefore, a kind of hierarchical fusion should be expected for capturing the complex correlations across modalities.

In this paper, to capture the complex correlations across modalities, we propose to densely integrate the representations of different networks, where the higher joint representation not only fuses the modality-specific representations in the same layer but also is conditioned on the lower joint one, which is named as \emph{Dense Multimodal Fusion} (DMF).
Different from the traditional fusion scheme based on deep networks, the learned joint representation in the hidden layer can simultaneously capture the covariation in the early fusion and the correlation between the inherent semantic of modalities. More importantly, the dense fusion provides multiple learning paths to enhance the interaction across modalities. For example, when one modality is with high uncertainty or missing, it can be efficiently inferred from the multi-level fused information.
To evaluate the proposed DMF scheme, we perform different multimodal tasks on several benchmark datasets, including AVSR, image-text classification, and cross-modal retrieval.
Extensive experiments show that DMF is superior to the traditional fusion schemes in these tasks, not only in the conditions of multimodal inputs but also unimodal input.


\section{Multimodal Deep Learning}

\begin{figure*}[t]
\centering
\includegraphics[width=13cm]{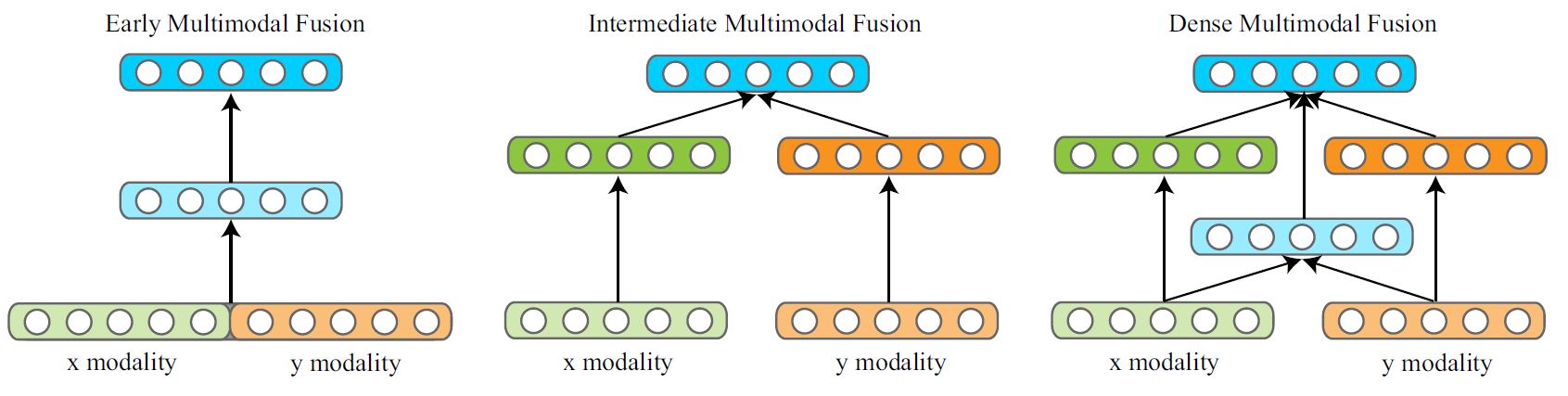}\\
\caption{Different multimodal fusion strategies based on deep networks: early fusion (left), intermediate fusion (middle), and the proposed dense fusion (right).}\label{ill}
\end{figure*}

To capture the correlation across modalities, an intuitive way is to directly concatenate the different features of them, then employ multiple layers of nonlinear transformation to generate the high-level joint representation~\cite{nojavanasghari2016deep}, which is named as \emph{Early Multimodal Fusion} (EMF), as shown in Fig.~\ref{ill}. When the deep network is viewed as a kind of discriminative model, such fusion scheme is truly based on the feature level.
Early fusion is easy to capture the covariation between modalities, or other correlations exist at the feature level, meanwhile, it is the simplest to implement~\cite{katsaggelos2015audiovisual}.
Unfortunately, although such fusion increases the dimensionality, it lacks the ability in capturing more complex correlation across modalities~\cite{ngiam2011multimodal}.
Another issue is that different features have different properties but need to be in the same space type, which requires to convert and scale them.

To tackle the problems of EMF, recent works propose to build a joint hidden layer based on the outputs of modality-specific networks. The general idea behind such fusion scheme is to reduce the influence of individual differences and improve the shared semantic~\cite{srivastava2012multimodal}.
The earliest representative work is MDAE~\cite{ngiam2011multimodal}. It is a AVSR network that learns a new shared representation layer across separated audio and visual networks, and employs it for multimodal reconstruction.
As the shared layer exists in the middle part of the whole multimodal network, such fusion is named as \emph{Intermediate Multimodal Fusion} (IMF)~\cite{ramachandram2017deep}, as shown in Fig.~\ref{ill}.
In view of the explainable rationale in semantic fusion, IMF has derived multiple variants in different multimodal task, such as multimodal deep belief network (MDBN) in AVSR~\cite{huang2013audio}, MDBM in image-text classification~\cite{srivastava2012multimodal}, \emph{Correspondence Autoencoder} (Corr-AE) in cross-modal retrieval \cite{feng2014cross, wang2014deeply}, deep fusion in affect prediction~\cite{nojavanasghari2016deep} and video description generation~\cite{jin2016video}, etc.  More related works can be found in the recent survey~\cite{baltruvsaitis2018multimodal}.

Although these IMF networks show promised results in different tasks, the fusion is just performed on one specific level, which therefore fails to capture the correlation in other layers, such as covariation~\cite{shivappa2010audiovisual}. Moreover, there exists only one path that connects two modalities, and such prolix inferring progress weakens the interaction and supervision across modalities.
Hence, the effective fusion at different levels is highly expected as opposed to the single fusion layer~\cite{ramachandram2017deep}.
In the unimodal scenario, Karpathy~\emph{et al.}~\cite{karpathy2014large} show that gradually fusing the video temporal information can provide the higher layers more global information in the video classification task. Du \emph{et al.}~\cite{du2015hierarchical} propose to hierarchically fuse the skeleton parts and gradually constitute the whole skeleton at the higher layers for action recognition.
While, there is few work that models the intricate multimodal correlation via densely fusing the representations at different depths.
Although Suk~\emph{et al.}~\cite{suk2014hierarchical} propose to perform the fusion based on patch-level and the image-level representation, the shared representation learning is only considered for patch feature learning via the standard MDBM~\cite{srivastava2012multimodal}, hence it is actually the same as conventional IMF scheme.
The most related multimodal work~\cite{neverova2016moddrop} is to progressively fuse different modalities (e.g., fusing gray scale and depth firstly, then combing the joint representation with audio/pose descriptor), hence it has not learned the hierarchical correlation between each two modalities.

Apart from multimodal learning, other tasks also hope to capture effective correlations between different networks. For example, action recognition aims to capture the correspondence between temporal and spatial network by resorting to arithmetical operation~\cite{park2016combining,feichtenhofer2016convolutional} and residual connection~\cite{feichtenhofer2016spatiotemporal} on pre-defined levels. But they fail to simultaneously capture the hierarchical correlation. More importantly, they do not share the same problem with multimodal learning, as we aim to establish reliable joint representation (via multiple shared layers) instead of the direct spatiotemporal correspondence.

\section{Dense Multimodal Fusion}
Based on the analysis about the previous multimodal networks, we can easily find that they share the same fusion scheme that consists of one shared layer and two modality-specific layers, i.e., the bottom two layers in EMF and the top two in IMF, as shown in Fig.~\ref{ill}. Such multimodal units have the ability in capturing the correlation between different input layers \cite{ngiam2011multimodal,sohn2014improved}.
To simultaneously capture the correlations in each layer, multiple shared layers should be considered.
In this paper, we employ dense multimodal fusion to learn the complex hierarchical correlations between the representations of different modalities, as shown in Fig.~\ref{ill}.

There are mainly two parts that constitute the proposed DMF, the stacked modality-specific layers and the stacked shared layers.
For the former, we can obtain multiple representations in different levels after several stacked layers of nonlinear transformation for each modality, which contain not only the detailed descriptions but also the semantic information.
For the latter, to capture the correlations between different modalities in each representation level, several shared layers are densely stacked to correlate modality-specific networks.
Each shared layer not only has the ability to capture the correlation in the current level, but also has the capacity to learn the dependency among the correlations.
Hence, DMF can capture more complex hierarchical correlation between modalities, and the experimental results also confirm this.

Actually, when there are only two multimodal units in the DMF (i.e., the example in Fig.~\ref{ill}), it can be viewed as a kind of combination of EMF and IMF.
Hence, DMF can simultaneously enjoy the merits of early fusion and intermediate fusion.
Moreover, when there are two more stacked multimodal units, DMF can model more correlations in the middle layers, which are neglected in EMF and IMF.
In the following sections, we will provide a detailed analysis about this.

\subsection{Multi-paths for Multimodal Learning}

\begin{figure*}[t]
\begin{minipage}[t]{0.5\linewidth}
\centering
\includegraphics[width=2.3in]{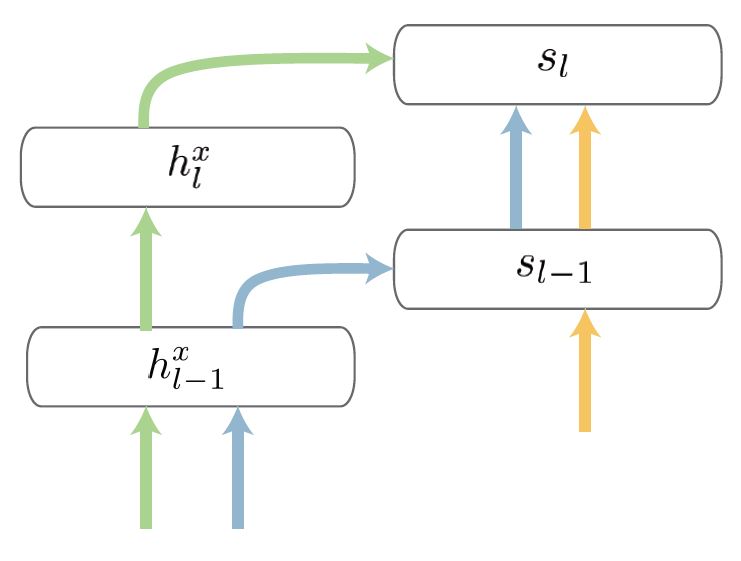}
\centerline{(a) Feed forward}
\label{feedforward}
\end{minipage}
\begin{minipage}[t]{0.5\linewidth}
\centering
\includegraphics[width=2.3in]{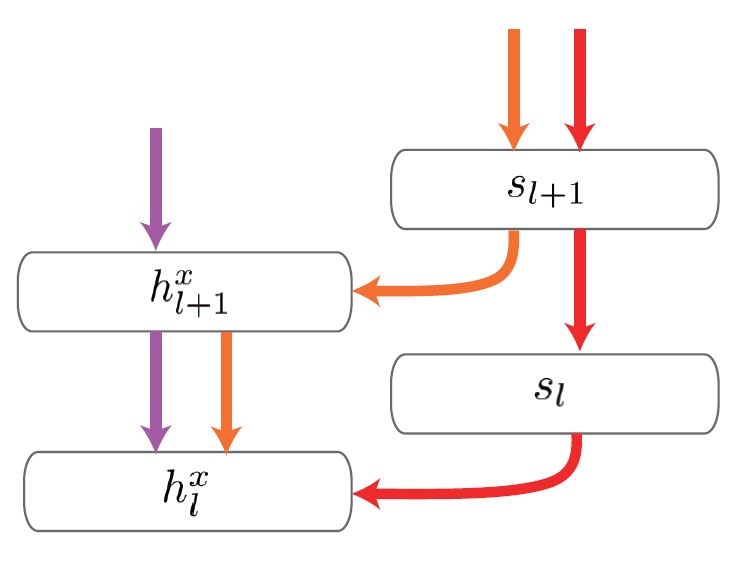}
\centerline{(b) Back propagation}
\label{backpropagation}
\end{minipage}
\caption{An illustration of feed forward and back propagate paths in DMF. Best viewed in color.}
\label{multipath}
\end{figure*}

To capture the complex hierarchical correlations between modalities, multiple learning paths (in both feed-forward and back-propagation) of the intra-modality and inter-modality should be expected.
However, the traditional multimodal networks, i.e., EMF and IMF, contain only one path for learning such correlations.
EMF is like a kind of unimodal network but based on the concatenated multimodal features, therefore the top layer just relies on the previous shared layer, and only the error of the shared layer can be propagated to each modality, which is weak in modeling the individual properties of each modality and the correlations in other levels.
Different from EMF, the feed-forward and back-propagation path of IMF rely on the modality-specific networks.
Concretely, for a IMF of $L$ layers, the top shared layer is obtained by the following equation\footnote{The modality layers and shared layer of one multimodal unit are deemed in the same layer.},
\begin{equation} \label{MPML_1}
{s_L} = {f}\left( {{W_{L}^{x \to s}}h_L^x + {W_{L}^{y \to s}}h_L^y} \right)
\end{equation}
where $f \left( \cdot \right)$ is the sigmoid activation function, ${W_{L}^{x \to s}}$ is the matrix of pairwise weights between elements of $h_L^x$and $s_L$, and similarly for ${W_{L}^{y \to s}}$. This is a standard multimodal unit. However, the hidden layers, $h_L^x$ and $h_L^y$, just rely on the modality-specific representations, hence, there exists only one path for learning the correlation across modalities. On the other hand, when backward  propagating the error, the gradient to current hidden layer of modality $x$ is written as,
\begin{equation} \label{MPML_2}
\frac{{\partial \varepsilon }}{{\partial h_l^x}} = \frac{{\partial \varepsilon }}{{\partial h_{l{\rm{ + }}1}^x}}\frac{{\partial h_{l{\rm{ + }}1}^x}}{{\partial h_l^x}},{\rm{~~~~~~~}}l = 1,2,...,L - 1
\end{equation}
where $\varepsilon$ stands for the objective function. The error of the joint representation is propagated via the term of ${{\partial \varepsilon } \mathord{\left/  {\vphantom {{\partial \varepsilon } {\partial h_{l + 1}^x}}} \right. \kern-\nulldelimiterspace} {\partial h_{l + 1}^x}}$, therefore, the modality-specific weights can be only optimized with respect to the correlation in the top layer.

Compared with the single learning path in EMF and IMF, DMF enjoys multiple paths when feeding the joint representation and propagating the errors, as shown in Fig. \ref{multipath}.
The corresponding update and optimization are written as,

\noindent{\emph{Feed-forward:}}
\begin{equation} \label{DMF_f}
{s_l} = {f}\left( {{W_{l}^{x \to s}}h_l^x + {W_{l}^{y \to s}}h_l^y + {W_{l-1}^{s}}s_{l-1} } \right),{\rm{~~~~~~~}}l = 2,...,L
\end{equation}
\begin{equation} \label{DMF_f}
h_l^x = f\left( {W_{l - 1}^xh_{l - 1}^x} \right),{\rm{~~~~~~~}}l = 2,...,L
\end{equation}

\noindent{\emph{Back-propagation:}}
\begin{equation} \label{dmn_b_1}
\frac{{\partial \varepsilon }}{{\partial h_l^x}} = \frac{{\partial \varepsilon }}{{\partial h_{l{\rm{ + }}1}^x}}\frac{{\partial h_{l{\rm{ + }}1}^x}}{{\partial h_l^x}} + \frac{{\partial \varepsilon }}{{\partial {s_l}}}\frac{{\partial {s_l}}}{{\partial h_l^x}},{\rm{~~~~~~~}}l = 1,2,...,L - 1
\end{equation}
\begin{equation} \label{dmn_b_2}
\frac{{\partial \varepsilon }}{{\partial {s_l}}} = \frac{{\partial \varepsilon }}{{\partial {s_{l + 1}}}}\frac{{\partial {s_{l + 1}}}}{{\partial {s_l}}},{\rm{~~~~~~~}}l = 1,2,...,L - 1
\end{equation}
where $W_{l - 1}^x$ is the modality-specific weight between layer $h_{l-1}^x$ and $h_{l}^x$, while ${W_{l-1}^{s}}$ is the weight between adjacent shared layers.
These weights of $\left\{W^x,~W^{x \to s},~W^s\right\}$ jointly model the intra- and inter-modalities correlations, similarly for modality $y$.
In Fig.~\ref{multipath}, we can easily find that there are three paths feeding the shared layer $s_l$ from modality $x$, where the green and yellow one are the same as the paths of IMF and EMF, respectively. These two help to capture the shallow and deep correlation between modalities. The remaining blue path indicates the correlations in the middle layers. When the number of stacked multimodal units increases, there will be more paths connected to higher shared layers.
Hence, DMF is more capable of capturing the complex correlation between modalities, not only the ones in the same layer but also in the cross-layers.

On the other hand, to efficiently optimize the network and infer the shared layers, multiple paths of back-propagation are performed in the DMF.
The purple and red path denote the error propagated from the modality-specific network and top shared layer, respectively. They preserve the consistency of intra-modality and inter-modality, which then help to establish the correlations in other layers.
Different from EMF and IMF, the distinct orange path is the error propagated via both the shared layer and modality layer.

As the shared layer in each level contains significant representation generated from both modalities, the orange path can provide efficient hierarchical supervision for each modality from the other one.
More specifically, let $M_{l}$ denote $\left( {{W_{l}^{x \to s}}h_l^x + {W_{l}^{y \to s}}h_l^y +  {W_{l-1}^{s}}s_{l-1} } \right)$, then the second term in Eq.~\ref{dmn_b_1} can be re-written into
\begin{equation} \label{DMF_b_1}
\frac{{\partial \varepsilon }}{{\partial {s_l}}}\frac{{\partial {s_l}}}{{\partial h_l^x}} = \left( {\frac{{\partial \varepsilon }}{{\partial {s_{l + 1}}}}\frac{{\partial {s_{l + 1}}}}{{\partial {s_l}}}} \right) \cdot \left( {W_l^{s \to x}f\left( {{M_l}} \right)\left( {1 - f\left( {{M_l}} \right)} \right)} \right),
\end{equation}
\begin{equation} \label{DMF_b_2}
\frac{{\partial {s_{l + 1}}}}{{\partial {s_l}}} = {\left( {W_l^s} \right)^T}f\left( {{M_{l + 1}}} \right)\left( {1 - f\left( {{M_{l + 1}}} \right)} \right).
\end{equation}
Recall that both the term of $M_{l}$ and $M_{l+1}$ contain the generated information of modality $y$, hence the error propagated to the current layer of $h_l^x$ contains hierarchical supervision from the other one.
Moreover, the multi-level cross-modal supervision can also come from the term of ${{\partial \varepsilon } \mathord{\left/ {\vphantom {{\partial \varepsilon } {\partial h_{l + 1}^x}}} \right. \kern-\nulldelimiterspace} {\partial h_{l + 1}^x}}$ in Eq.~\ref{dmn_b_1}. Hence, when one modality is damaged or missing, DMF can still have the ability to provide efficient supervision from the other one and learn effective joint representation, which is also confirmed in the experiments.



\begin{figure*}[t]
\centering
\includegraphics[width=11cm]{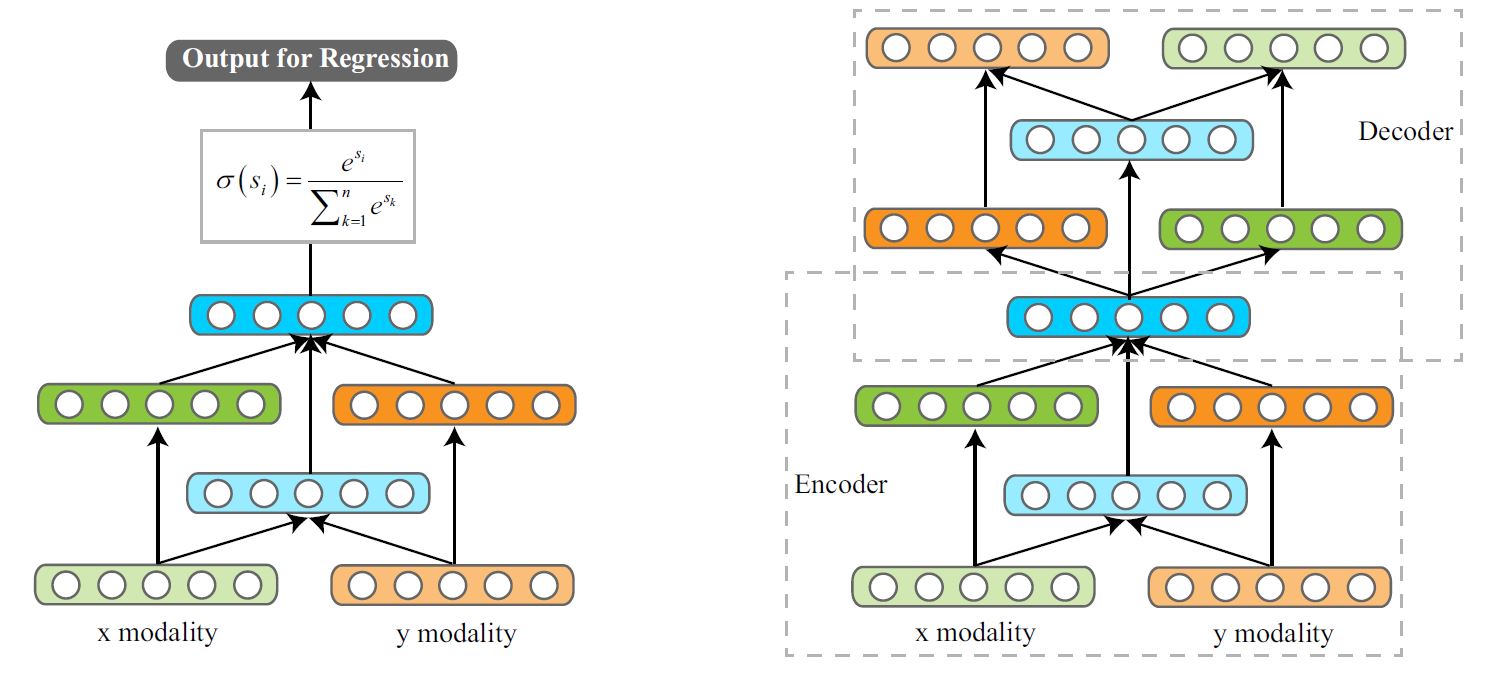}\\
\caption{The DMF variants in the discriminative (left) and generative (right) task, respectively.}\label{models}
\end{figure*}

\subsection{Model Variants}
Dense fusion is not one specific network architecture but a novel fusion scheme or mechanism. Hence, it has different model variants for different multimodal learning tasks.
One common task is taking advantage of the more valuable information of multiple modalities to perform more exact classification.
For this task, DMF can be viewed as a discriminative model, where a regression layer is performed over the top joint representation, as shown in Fig.~\ref{models}.
Such model can be finetuned to minimize the categorical cross-entropy after initializing the feed forward network.
Another common task is to infer the robust joint representation with different input modalities, which could be then used for cross-modal retrieval~\cite{wang2016comprehensive} or recognition~\cite{hu2016temporal}.
The cross-modal retrieval task takes the representation of one modality as the query to retrieve relevant items of another modality~\cite{li2017deep}.
To simultaneously preserve the inter- and intra-modal consistency under the unsupervised fashion, we propose to reconstruct the modalities by given the joint representation, which actually treats DMF as an ``encoder'' and the reversed one as a ``decoder'', as shown in Fig.~\ref{models}.
When there is only one modality available in the retrieval phase, DMF has to encode the joint representation based on single modality input and decode the representation into different modalities.
Such ``encoder-decoder'' network can also be used for recognition. We can take the encoded joint representation as the inputs to classifier (e.g., SVM). Such operation has been widely used in the traditional multimodal network.
As for the optimization of these DMF model variants, we initialize each multimodal unit with the Contrastive Divergence~\cite{hinton2002training} in the layer-wise fashion, then fine-tune the whole network based on the respective supervision above.


\section{Experiments}
\begin{figure*}[t]
\centering
\includegraphics[width=12cm]{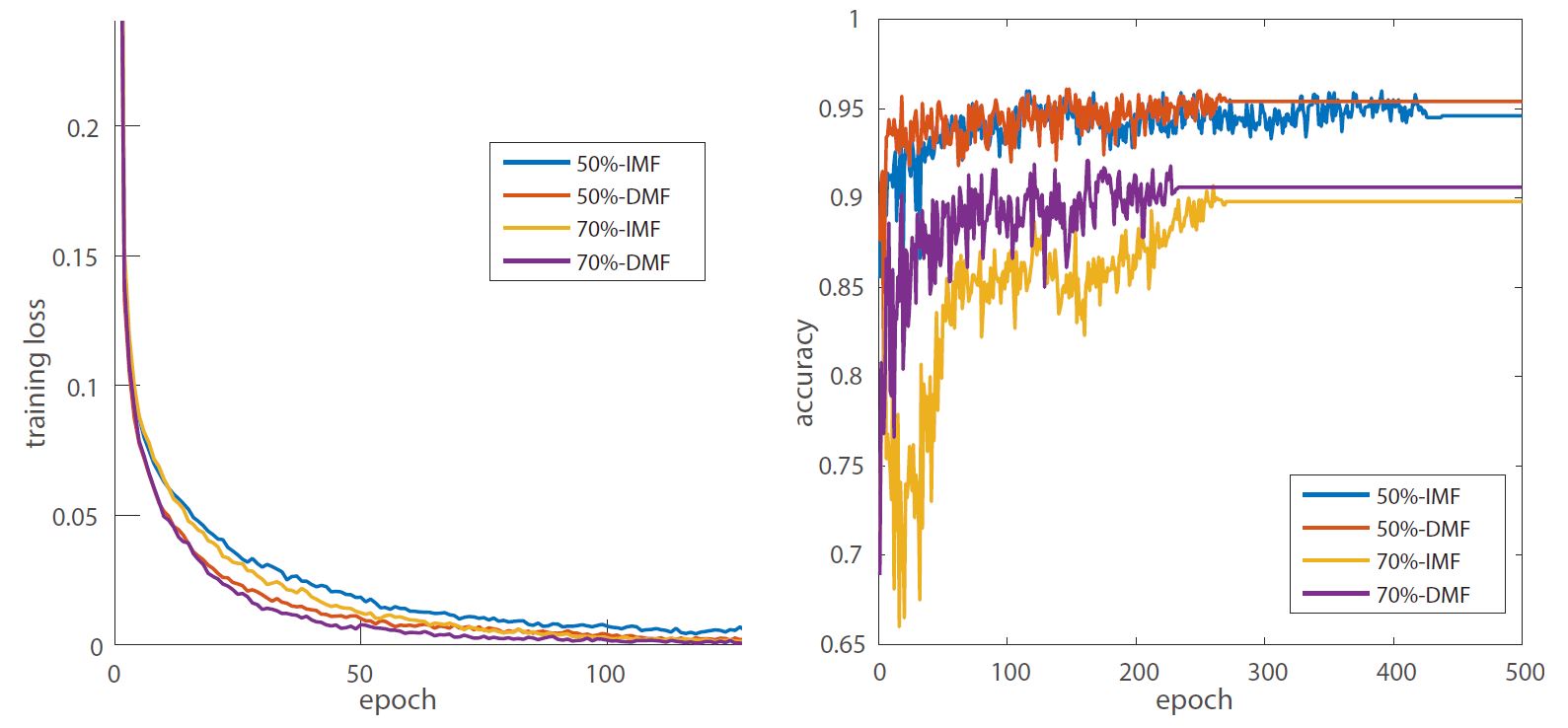}\\
\caption{The training loss and testing accuracy of DMF and IMF on the mnist dataset. The right half of digital image is manually damaged at different levels. Best viewed in color.}\label{mnist}
\end{figure*}

\subsection{Toy Example}
In this section, following~\cite{sohn2014improved,hu2016multimodal}, we first evaluate different fusion strategies (i.e., EMF, IMF, and DMF) on the MNIST dataset~\cite{lecun1998gradient}.
The MNIST dataset consists of 60,000 training and 10,000 testing images. These images of 28$\times$28 pixels are equally divided into right and left halves which are considered as two modalities describing the same digits~\cite{sohn2014improved}. To evaluate the ability in learning robust representation when faced with modalities with high uncertainty, we randomly set part of the right half to zeros at different levels, i.e., $\left\{ {0,~30\%,~50\%,~70\%} \right\}$.  And we also compare these methods under different input conditions, i.e., left+right and right modality.

Table.~\ref{Table1} shows the comparison results among EMF, IMF, and DMF.
The network for each image pathway is [392, 512, 128] and the shared pathway is [512, 256, 64]. Although EMF and IMF share the same unimodal paths as DMF, the additional shared pathway still leads to the growth of variable number (about 2-3 times larger). The increased variables are correlation learning-related, which means DMF has greater potential in capturing the complex multimodal correlation beyond traditional multimodal network. Hence, DMF does not suffer from the intractable overfitting problem, but significantly outperforms the other two fusion strategies in different input conditions.

Surprisingly, training much more parameters in DMF do not cost more time but less (similar in the following multimodal tasks), compared with classical fusion network, as shown in Fig.~\ref{mnist}.
The noticeable performance comes from the multiple efficient learning paths of DMF. Specifically, each unimodal layer can receive multiple kinds of error propagated from different layers (Eq.~\ref{dmn_b_1}) instead of the single error path of IMF (Eq.~\ref{MPML_2}). Moreover, these learning paths also contribute to the efficient hierarchical supervision in DMF (Eq.~\ref{DMF_b_1} and Eq.~\ref{DMF_b_2}). When one modality is badly damaged, the other reliable one can provide supervision information on different levels for it, while EMF and IMF only provide such information on the bottom and top layer, respectively. Hence, DMF still shows noticeable performance when we destroy one modality. These properties jointly contribute to the lower training loss, higher testing accuracy, and faster convergence of DMF.

\begin{table}[t]
  \centering
  \caption{Recognition performance (in error) for handwritten digit recognition on the MNIST dataset. The percentages indicate the degrees of damage to the right half of images. \label{Table1}}
  \renewcommand\arraystretch{1}
  \begin{tabular}{cccccc}
    \toprule
    Modality &  Models       &  0  &  30\%  &  50\% & 70\%  \\
    \midrule
    \multirow{3}*{Left+Right}   & EMF        & 1.90 &   2.60  & 4.70   & 12.20 \\
                                                      & IMF          & 1.60  &  2.40  &   5.40 &  10.20 \\
                                                      &  DMF           & \textbf{1.40}  &  \textbf{2.30}  &   \textbf{4.60} &   \textbf{9.40} \\
    \midrule
    \multirow{3}*{Right}           &  EMF        & 51.00  & 53.50   &  63.40 & 70.30 \\
                                                     & IMF          & 51.30  &  54.30  & 63.30 & 82.90  \\
                                                     &  DMF          & \textbf{39.90}   &  \textbf{47.60}  & \textbf{57.10} & \textbf{67.60}  \\
    \bottomrule
  \end{tabular}
\end{table}

\subsection{Audiovisual Speech Recognition}
Audiovisual speech recognition is a classical task that makes use of the information from audio and visual modality to perform robust speech recognition.
In this section, we compare DMF with  Active Appearance Model (AAM)~\cite{bear2015speaker}, MDAE~\cite{ngiam2011multimodal}, MDBN~\cite{huang2013audio}, \emph{Recurrent Temporal Multimodal RBM} (RTMRBM)~\cite{hu2016temporal}, \emph{Conditional RBM} (CRBM)~\cite{amer2014multimodal}, and CorrRNN~\cite{yang2017deep}. To be fair, we use the same discriminative model as MDAE and MDBN that employ the ``encoder-decoder'' framework and use SVM as the classifier.
The same network architecture is also adopted except the shared pathway, which is [1024, 512, 256].
The experiments are conducted on the AVLetters2 dataset~\cite{cox2008challenge}. It is about reading letters from A to Z, spoken by five people, seven times for each letter.
Similar with \cite{hu2016temporal}, we use the letters spoken by four people for training and the rest for testing.
Both unimodal and multimodal inputs are considered, where 4 audio frames and 1 video frame are used as the multimodal inputs to the networks.

\begin{wrapfigure}{R}{0.5\textwidth}
  \centering
    \caption{The mean accuracy of speech recognition on AVLetters2. All the models are evaluated with different input modalities. For the unimodal input, one modality is preserved while the other one is set to
zero. \label{Table2}}
\vskip 0.15in
  \begin{tabular}{p{1.5cm}<{\centering}p{0.8cm}<{\centering}p{0.8cm}<{\centering}p{0.8cm}<{\centering}}
    \toprule
    Modality       &  A &  V &  A+V  \\
    \midrule
    AAM         &     15.2    &     -     &    -      \\
    MDAE        &     -    &     -     &     67.89        \\
    MDBN        &     -    &     -     &    54.1       \\
    CRBM      & -        & -           & 74.08 \\
    RTMRBM  &  75.85 &  31.21 & 74.77\\
    CorrRNN   &   81.36     &    60.17   &   76.32 \\
    DMF  &  \textbf{86.43} &  \textbf{75.87} & \textbf{80.43}\\
    \bottomrule
  \end{tabular}
\end{wrapfigure}


Table~\ref{Table2} shows the results in accuracy.
We can find that DMF shows significant improvement over the other ones. Note that RTMRBM, CRBM, and CorrRNN are temporal models that can capture the dependence among the multimodal sequence.
However, DMF is not a temporal model but still outperforms these methods, which shows its ability in learning effective joint representation.
Moreover, when only the visual modality is available, DMF has a noticeable improvement. This is because DMF can establish efficient correlation between modalities in each layer, which helps to make one modality learn from the other one. Even so, the visual modality still lowers the performance of multimodal inputs to some extent, but it is a common situation~\cite{hu2016temporal}.


\subsection{Cross-modal Retrieval}
In this experiment, we focus on two cross-modal retrieval tasks, i.e., image2text (I2T) and text2image (T2I).
We compare our model with four unsupervised methods, including CCA~\cite{rasiwasia2010new}, CMFH (without binary constraint)~\cite{ding2016large}, LCFS~\cite{wang2013learning}, and Corr-Full-AE~\cite{feng2014cross}.
The benchmark image-text dataset of Wiki  is chosen for evaluation~\cite{rasiwasia2010new}.  For each pair, the image modality is represented as 128-D SIFT descriptor histograms, and text is expressed as 10-D semantic vector. These pairs are annotated with one of 10 topic labels. In this paper, we choose 25\% of the dataset as the query set and the rest for retrieval set. And we still use the ``encoder-decoder'' framework and reconstruct both modalities based on the query modality. 

We show the ranking performance in Table~\ref{Table3}. It is obvious that DMF enjoys the best results among these methods. Specifically, the non-linear methods outperform the linear ones, which illustrates that they are better at capturing the complex non-linear correlation across modalities. On the other hand, Corr-Full-AE is similar as IMF, which attempts to capture the correlation between the middle layers of modality-specific networks.  However, it aims to minimize the differences between the representations instead of learning the joint representation. Such framework makes it difficult to train and optimize.
In contrast, DMF is easier to optimize and shows better performance.

\begin{table}[h]
  \centering
    \caption{The ranking performance of cross-modal retrieval on Wiki dataset.  \label{Table3}}
  \renewcommand\arraystretch{1}
  \begin{tabular}{cccccc}
    \toprule
    mAP       &  CCA &  CMFH  & LCFS &  Corr-Full-AE & DMF \\
    \midrule
    I2T              &  0.2490 &  0.2551   &   0.2798  &   0.2634       &   \textbf{0.2921}\\
    T2I              &  0.1960 &   0.5407  & 0.2141    &   0.5418      &     \textbf{0.5612}\\
    \bottomrule
  \end{tabular}
\end{table}

\subsection{Multimodal Classification}
For multimodal classification, we compare with different variants of IMF, including MDAE~\cite{ngiam2011multimodal}, MDBN~\cite{huang2013audio}, MDBM~\cite{srivastava2012multimodal}, and MinVI~\cite{sohn2014improved}, where these models take the same number of layers and units as DMF except the dense shared layers.
We employ DMF to reconstruct both modalities based on the multimodal inputs, and use the top joint representation as the input features to the same discriminative model (SVM) as other models.
MIR-FLICKR dataset~\cite{huiskes2008mir} is employed for evaluation, which consists of 1 million images and corresponding user tags collected from the photography website Flickr.
We use the same visual and text features as~\cite{srivastava2012multimodal}, where the image feature is pre-processed into zero-mean and unit variance for each dimension and the text input is represented by a word count of the 2000 most frequency tags. We randomly select 15,000 for training and the rest 10,000 for testing.

Table~\ref{Table4} shows the comparison results. DMF outperforms other IMF models without additional fine-tuning.
MDBM is an undirect multimodal model, where the responsibility of the multimodal modeling is spread over the entire network. Hence it has the ability to indirectly capture the correlation in the lower layers and is slightly better than MDBN and MDAE but worse than DMF. DMF can directly learn the correlation in each layer as well as the dependency between shared layers, hence it shows the best performance among these methods. On the other hand, MinVI aims to reduce the variant information across modalities, which is good at inferring mutual information but weak in learning the private properties of each modality. However, DMF can simultaneously preserve the mutual and private information via the dense shared layers and modality-specific layers.


\begin{table}[h]
  \centering
  \caption{Classification results on Flickr dataset. MDRNN$\dag$ is the same as MinVI but fine-tuned by a second-stage multimodal RNN. \label{Table4}}
  \renewcommand\arraystretch{1}
  \begin{tabular}{p{1cm}<{\centering}p{0.8cm}<{\centering}p{0.8cm}<{\centering}p{0.8cm}<{\centering}p{0.8cm}<{\centering}p{0.8cm}<{\centering}|p{1.3cm}<{\centering}}
    \toprule
    Models       &  MDAE &  MDBN  &  MDBM & MinVI& DMF  & MDRNN$\dag$  \\
    \midrule
    mAP   &  0.600 &  0.599   &  0.609 & 0.574  & \textbf{0.618}& 0.686    \\
    \bottomrule
  \end{tabular}
\end{table}

\section{Conclusion}
Deep model can provide representations in different levels for single modality, such as raw features in the bottom and abstract semantic in the intermediate layer.
Hence, there exist multiple kinds of hierarchical correlations between different modality-specific networks, which have not been effectively explored so far.
In this paper, we propose to densely correlate these representations of different modalities layer-by-layer, where the shared layer not only models the correlation in the current level but also depends on the lower one.
Such dense fusion not only rewards it the advantages of early and intermediate fusion multimodal network but also the multiple learning paths that help to capture more complex correlation and accelerate convergence.
Moreover, DMF can also provide hierarchical cross-modal supervision for the modality with high uncertainty.

\bibliographystyle{ieee}
\bibliography{smn}

\begin{thebibliography}{10}\itemsep=-1pt

\bibitem{amer2014multimodal}
M.~R. Amer, B.~Siddiquie, S.~Khan, A.~Divakaran, and H.~Sawhney.
\newblock Multimodal fusion using dynamic hybrid models.
\newblock In {\em Applications of Computer Vision (WACV), 2014 IEEE Winter
  Conference on}, pages 556--563. IEEE, 2014.

\bibitem{baltruvsaitis2018multimodal}
T.~Baltru{\v{s}}aitis, C.~Ahuja, and L.-P. Morency.
\newblock Multimodal machine learning: A survey and taxonomy.
\newblock {\em IEEE Transactions on Pattern Analysis and Machine Intelligence},
  2018.

\bibitem{bear2015speaker}
H.~L. Bear, S.~J. Cox, and R.~W. Harvey.
\newblock Speaker-independent machine lip-reading with speaker-dependent viseme
  classifiers.
\newblock In {\em AVSP}, pages 190--195, 2015.

\bibitem{cox2008challenge}
S.~J. Cox, R.~W. Harvey, Y.~Lan, J.~L. Newman, and B.-J. Theobald.
\newblock The challenge of multispeaker lip-reading.
\newblock In {\em AVSP}, pages 179--184, 2008.

\bibitem{ding2016large}
G.~Ding, Y.~Guo, J.~Zhou, and Y.~Gao.
\newblock Large-scale cross-modality search via collective matrix factorization
  hashing.
\newblock {\em IEEE Transactions on Image Processing}, 25(11):5427--5440, 2016.

\bibitem{du2015hierarchical}
Y.~Du, W.~Wang, and L.~Wang.
\newblock Hierarchical recurrent neural network for skeleton based action
  recognition.
\newblock In {\em Proceedings of the IEEE conference on computer vision and
  pattern recognition}, pages 1110--1118, 2015.

\bibitem{feichtenhofer2016spatiotemporal}
C.~Feichtenhofer, A.~Pinz, and R.~Wildes.
\newblock Spatiotemporal residual networks for video action recognition.
\newblock In {\em Advances in Neural Information Processing Systems}, pages
  3468--3476, 2016.

\bibitem{feichtenhofer2016convolutional}
C.~Feichtenhofer, A.~Pinz, and A.~Zisserman.
\newblock Convolutional two-stream network fusion for video action recognition.
\newblock In {\em Proceedings of the IEEE Conference on Computer Vision and
  Pattern Recognition}, pages 1933--1941, 2016.

\bibitem{feng2014cross}
F.~Feng, X.~Wang, and R.~Li.
\newblock Cross-modal retrieval with correspondence autoencoder.
\newblock In {\em Proceedings of the 22nd ACM international conference on
  Multimedia}, pages 7--16. ACM, 2014.

\bibitem{hinton2002training}
G.~E. Hinton.
\newblock Training products of experts by minimizing contrastive divergence.
\newblock {\em Neural computation}, 14(8):1771--1800, 2002.

\bibitem{holmes2005multisensory}
N.~P. Holmes and C.~Spence.
\newblock Multisensory integration: space, time and superadditivity.
\newblock {\em Current Biology}, 15(18):R762--R764, 2005.

\bibitem{hu2016temporal}
D.~Hu, X.~Li, et~al.
\newblock Temporal multimodal learning in audiovisual speech recognition.
\newblock In {\em Proceedings of the IEEE Conference on Computer Vision and
  Pattern Recognition}, pages 3574--3582, 2016.

\bibitem{hu2016multimodal}
D.~Hu, X.~Lu, and X.~Li.
\newblock Multimodal learning via exploring deep semantic similarity.
\newblock In {\em Proceedings of the 2016 ACM on Multimedia Conference}, pages
  342--346. ACM, 2016.

\bibitem{huang2013audio}
J.~Huang and B.~Kingsbury.
\newblock Audio-visual deep learning for noise robust speech recognition.
\newblock In {\em Acoustics, Speech and Signal Processing (ICASSP), 2013 IEEE
  International Conference on}, pages 7596--7599. IEEE, 2013.

\bibitem{huiskes2008mir}
M.~J. Huiskes and M.~S. Lew.
\newblock The mir flickr retrieval evaluation.
\newblock In {\em Proceedings of the 1st ACM international conference on
  Multimedia information retrieval}, pages 39--43. ACM, 2008.

\bibitem{jin2016video}
Q.~Jin and J.~Liang.
\newblock Video description generation using audio and visual cues.
\newblock In {\em Proceedings of the 2016 ACM on International Conference on
  Multimedia Retrieval}, pages 239--242. ACM, 2016.

\bibitem{karpathy2014large}
A.~Karpathy, G.~Toderici, S.~Shetty, T.~Leung, R.~Sukthankar, and L.~Fei-Fei.
\newblock Large-scale video classification with convolutional neural networks.
\newblock In {\em Proceedings of the IEEE conference on Computer Vision and
  Pattern Recognition}, pages 1725--1732, 2014.

\bibitem{katsaggelos2015audiovisual}
A.~K. Katsaggelos, S.~Bahaadini, and R.~Molina.
\newblock Audiovisual fusion: Challenges and new approaches.
\newblock {\em Proceedings of the IEEE}, 103(9):1635--1653, 2015.

\bibitem{lecun1998gradient}
Y.~LeCun, L.~Bottou, Y.~Bengio, and P.~Haffner.
\newblock Gradient-based learning applied to document recognition.
\newblock {\em Proceedings of the IEEE}, 86(11):2278--2324, 1998.

\bibitem{li2017deep}
X.~Li, D.~Hu, and F.~Nie.
\newblock Deep binary reconstruction for cross-modal hashing.
\newblock {\em arXiv preprint arXiv:1708.05127}, 2017.

\bibitem{neverova2016moddrop}
N.~Neverova, C.~Wolf, G.~Taylor, and F.~Nebout.
\newblock Moddrop: adaptive multi-modal gesture recognition.
\newblock {\em IEEE Transactions on Pattern Analysis and Machine Intelligence},
  38(8):1692--1706, 2016.

\bibitem{ngiam2011multimodal}
J.~Ngiam, A.~Khosla, M.~Kim, J.~Nam, H.~Lee, and A.~Y. Ng.
\newblock Multimodal deep learning.
\newblock In {\em Proceedings of the 28th international conference on machine
  learning (ICML-11)}, pages 689--696, 2011.

\bibitem{nojavanasghari2016deep}
B.~Nojavanasghari, D.~Gopinath, J.~Koushik, T.~Baltru{\v{s}}aitis, and L.-P.
  Morency.
\newblock Deep multimodal fusion for persuasiveness prediction.
\newblock In {\em Proceedings of the 18th ACM International Conference on
  Multimodal Interaction}, pages 284--288. ACM, 2016.

\bibitem{park2016combining}
E.~Park, X.~Han, T.~L. Berg, and A.~C. Berg.
\newblock Combining multiple sources of knowledge in deep cnns for action
  recognition.
\newblock In {\em Applications of Computer Vision (WACV), 2016 IEEE Winter
  Conference on}, pages 1--8. IEEE, 2016.

\bibitem{potamianos2004audio}
G.~Potamianos, C.~Neti, J.~Luettin, and I.~Matthews.
\newblock Audio-visual automatic speech recognition: An overview.
\newblock {\em Issues in visual and audio-visual speech processing}, 22:23,
  2004.

\bibitem{ramachandram2017deep}
D.~Ramachandram and G.~W. Taylor.
\newblock Deep multimodal learning: A survey on recent advances and trends.
\newblock {\em IEEE Signal Processing Magazine}, 34(6):96--108, 2017.

\bibitem{rasiwasia2010new}
N.~Rasiwasia, J.~Costa~Pereira, E.~Coviello, G.~Doyle, G.~R. Lanckriet,
  R.~Levy, and N.~Vasconcelos.
\newblock A new approach to cross-modal multimedia retrieval.
\newblock In {\em Proceedings of the 18th ACM international conference on
  Multimedia}, pages 251--260. ACM, 2010.

\bibitem{salakhutdinov2009deep}
R.~Salakhutdinov and G.~Hinton.
\newblock Deep boltzmann machines.
\newblock In {\em Artificial Intelligence and Statistics}, pages 448--455,
  2009.

\bibitem{salakhutdinov2009semantic}
R.~Salakhutdinov and G.~Hinton.
\newblock Semantic hashing.
\newblock {\em International Journal of Approximate Reasoning}, 50(7):969--978,
  2009.

\bibitem{shivappa2010audiovisual}
S.~T. Shivappa, M.~M. Trivedi, and B.~D. Rao.
\newblock Audiovisual information fusion in human--computer interfaces and
  intelligent environments: A survey.
\newblock {\em Proceedings of the IEEE}, 98(10):1692--1715, 2010.

\bibitem{snoek2005early}
C.~G. Snoek, M.~Worring, and A.~W. Smeulders.
\newblock Early versus late fusion in semantic video analysis.
\newblock In {\em Proceedings of the 13th annual ACM international conference
  on Multimedia}, pages 399--402. ACM, 2005.

\bibitem{sohn2014improved}
K.~Sohn, W.~Shang, and H.~Lee.
\newblock Improved multimodal deep learning with variation of information.
\newblock In {\em Advances in Neural Information Processing Systems}, pages
  2141--2149, 2014.

\bibitem{srivastava2012multimodal}
N.~Srivastava and R.~R. Salakhutdinov.
\newblock Multimodal learning with deep boltzmann machines.
\newblock In {\em Advances in neural information processing systems}, pages
  2222--2230, 2012.

\bibitem{stein1993merging}
B.~E. Stein and M.~A. Meredith.
\newblock {\em The merging of the senses.}
\newblock The MIT Press, 1993.

\bibitem{suk2014hierarchical}
H.-I. Suk, S.-W. Lee, D.~Shen, A.~D.~N. Initiative, et~al.
\newblock Hierarchical feature representation and multimodal fusion with deep
  learning for ad/mci diagnosis.
\newblock {\em NeuroImage}, 101:569--582, 2014.

\bibitem{wang2013learning}
K.~Wang, R.~He, W.~Wang, L.~Wang, and T.~Tan.
\newblock Learning coupled feature spaces for cross-modal matching.
\newblock In {\em Computer Vision (ICCV), 2013 IEEE International Conference
  on}, pages 2088--2095. IEEE, 2013.

\bibitem{wang2016comprehensive}
K.~Wang, Q.~Yin, W.~Wang, S.~Wu, and L.~Wang.
\newblock A comprehensive survey on cross-modal retrieval.
\newblock {\em arXiv preprint arXiv:1607.06215}, 2016.

\bibitem{wang2014deeply}
W.~Wang, Z.~Cui, H.~Chang, S.~Shan, and X.~Chen.
\newblock Deeply coupled auto-encoder networks for cross-view classification.
\newblock {\em arXiv preprint arXiv:1402.2031}, 2014.

\bibitem{yang2017deep}
X.~Yang, P.~Ramesh, R.~Chitta, S.~Madhvanath, E.~A. Bernal, and J.~Luo.
\newblock Deep multimodal representation learning from temporal data.
\newblock {\em CoRR abs/1704.03152}, 2017.

\end{thebibliography}

\end{document}